\documentclass[pre,aps,twocolumn,amsmath,amssymb]{revtex4-1}

\pdfoutput=1
\usepackage{hyperref}
\usepackage{graphicx}
\usepackage[usenames]{color}
\usepackage{bm}
\usepackage{hyperref}
\usepackage{url}
\usepackage{float}

\def\cal#1{\mathcal{#1}}
\def\eqq#1{Eq.~(\ref{#1})}
\def\eq#1{(\ref{#1})}

\def\f#1{Fig.~\ref{#1}}

\def\c#1{~\cite{#1}}
\def\cc#1{Ref.\c{#1}}

\def\s#1{Section~\ref{#1}}

\def\o{{\cal O}}

\def\beq{\begin{equation}}
\def\eeq{\end{equation}}
\def\bea{\begin{eqnarray}}
\def\eea{\end{eqnarray}}

\begin{document}

\title{Improving the accuracy of nearest-neighbor classification using principled construction and stochastic sampling of training-set centroids}
\author{Stephen Whitelam}\email{swhitelam@lbl.gov}
\affiliation{Molecular Foundry, Lawrence Berkeley National Laboratory, 1 Cyclotron Road, Berkeley, CA 94720, USA}
\begin{abstract}
A conceptually simple way to classify images is to directly compare test-set data and training-set data. The accuracy of this approach is limited by the method of comparison used, and by the extent to which the training-set data cover configuration space. Here we show that this coverage can be substantially increased using coarse graining (replacing groups of images by their centroids) and stochastic sampling (using distinct sets of centroids in combination). We use the MNIST and Fashion-MNIST data sets to show that a principled coarse-graining algorithm can convert training images into fewer image centroids without loss of accuracy of classification of test-set images by nearest-neighbor classification. Distinct batches of centroids can be used in combination as a means of stochastically sampling configuration space, and can classify test-set data more accurately than can the unaltered training set. On the MNIST and Fashion-MNIST data sets this approach converts nearest-neighbor classification from a mid-ranking- to an upper-ranking member of the set of classical machine-learning techniques.

\end{abstract}
\maketitle
  
\section{Introduction}

Machine learning, used for many years in fields such as image recognition\c{lecun2015deep,nasrabadi2007pattern,lecun1990handwritten,hinton1997modeling} and game playing\c{quinlan1983learning, samuel1959some}, is becoming established in the physical sciences\c{ferguson2018machine}. Machine learning has been used to calculate thermodynamic quantities of molecular systems, both classical\c{desgranges2018new,portman2017sampling} asnd quantum\c{mills2017deep,artrith2018constructing,singraber2018density}, and to predict and guide the outcome of nonequilibrium processes such as self-assembly\c{thurston2018machine,whitelam2020learning}. It is also natural to expect physical concepts, and methods developed in the physical sciences, to be used increasingly to study and improve machine-learning tools\c{kossio2018growing}. Here we show that physics-inspired ideas of coarse graining and sampling can be used to improve the efficiency and accuracy of nearest-neighbor image classification.

A conceptually simple method of image recognition is to assign to a test-set image the type of the most similar training-set image\c{zhang2007ml,bhatia2010survey,fritzke1995growing,nova2014review,marsland2002self}. The accuracy of such an approach is limited by the method of image comparison, and by the extent to which the training-set data ``sample'' configuration space. In \f{fig0}(a) we show 36 examples of 3s and 5s taken from the MNIST data set, which consists of hand-drawn digits\c{lecun1998gradient}, to emphasize that a single concept can be represented by many different configurations. One way to better sample configuration space is to synthetically expand the set of training data, e.g. by effecting translations or elastic distortions of digits\c{lecun1998gradient}. One drawback of such approaches is that they require prior knowledge of the concept to be classified. 

Here we use the MNIST and Fashion-MNIST\c{xiao2017fashion} data sets to show that principled coarse graining and sampling can be used to substantially reduce error rates of memory-based, nearest-neighbor classifier schemes without requiring information beyond that contained in the training set. 

We show how to identify a set of coarse-grained symbols or {\em memories} able to correctly classify training-set data by capturing the diversity of the set of symbols while omitting redundant objects [see \f{fig0}(b)]. In this context ``coarse graining'' means combining symbols into groups and identifying their centroids, in order to produce a sparse description of configuration space; it does not mean sampling images at lower resolution. The process of coarse graining is similar in spirit to the process enacted by particle-clustering algorithms\c{wolff1989collective,liu2004rejection,whitelam2011approximating}; by a supervised $k$-means algorithm\c{wagstaff2001constrained}; and by learning vector quantization classifiers\c{nova2014review} such as the growing neural gas\c{fritzke1995growing} or growing self-organizing network\c{marsland2002self}. For MNIST or Fashion-MNIST data, a set of $N$ training-set symbols can be described by a set of about $0.15 N$ or $0.2N$ memories, respectively. Memories classify unseen test-set symbols, via nearest-neighbor classification, as accurately or slightly more accurately than do the symbols from which they are derived; a natural expectation is the opposite\c{lecun1998gradient}. 

Moreover, memory sets created by drawing batches of symbols stochastically from the training set can be used to classify test-set data substantially more accurately than can the original training set. In effect, memories can be used to ``sample'' configuration space. The idea is similar in principle to combining nearest-neighbor classifiers into voting blocs\c{skalak1997prototype}; however, here we perform classification using objects not present in the original training set, and use the process of sampling to identify new symbols that better describe unseen data. 

Nearest-neighbor classification using the normalized vector dot product applied to raw pixels classifies the MNIST and Fashion-MNIST data sets at an error rate of about $2.8\%$ and $15\%$ respectively, making them mid-ranking members of the set of classical machine-learning methods\c{xiao2017fashion}. We show here that under coarse graining and sampling these error rates decrease to about $1.6\%$ and $10\%$ respectively, comparable with the best classical machine-learning methods surveyed in \cc{xiao2017fashion}. Deep-learning methods are generally more accurate than classical ones\c{fmn}, but the point of this paper is to show the improvement possible using principled combination and sampling of training-set data, without knowledge of the concept being classified and without knowing that the data correspond to images (and so not allowing expansion of the training set using e.g. elastic distortions or translations).

In what follows we describe the method using the MNIST data set as an illustration, and apply it to MNIST and Fashion-MNIST.

\begin{figure}[] 
   \includegraphics[width=\linewidth]{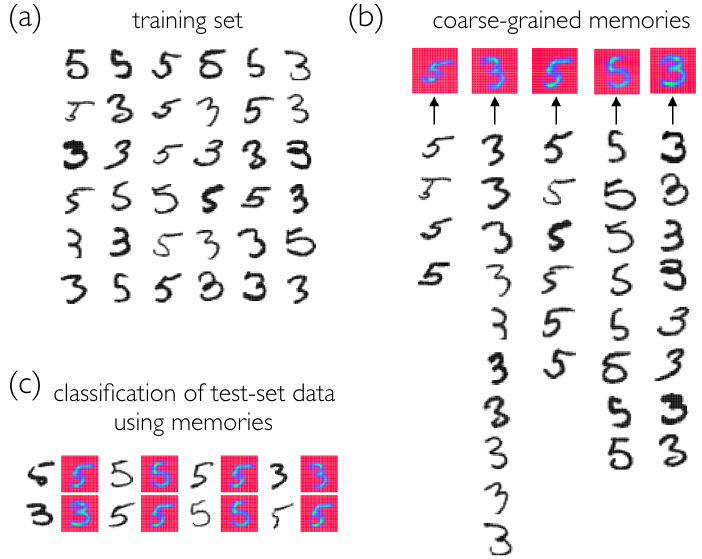} 
   \caption{(a) A set of 36 examples taken from the MNIST training set gives rise to 5 memories [panel (b)] upon coarse graining according to the procedure described in this paper. (b) Each of the 36 symbols is ``stored'' in a memory; memories are the centroids of their constituent symbols. Memories are colored so that red and green areas correspond to white and black portions of symbols, respectively. These 5 memories correctly classify the 36 symbols by nearest-neighbor classification: the coarse-graining algorithm ``learns'' to partition configuration space accurately. (c) Memories can be used to classify unseen test-set symbols as accurately as can the symbols from which they are derived: here we show 8 correctly classified test-set digits next to the memory they most closely resemble. As we show in this paper, combinations of memories derived from different symbol sets can be used to ``sample'' configuration space and achieve more accurate classification, of unseen symbols from the test set, than can the original training-set symbols.}
   \label{fig0}
\end{figure}

\section{Image recognition via nearest-neighbor classification}
\subsection{Unaltered training set}
\begin{figure*}[t] 
   \centering
   \includegraphics[width=0.8\linewidth]{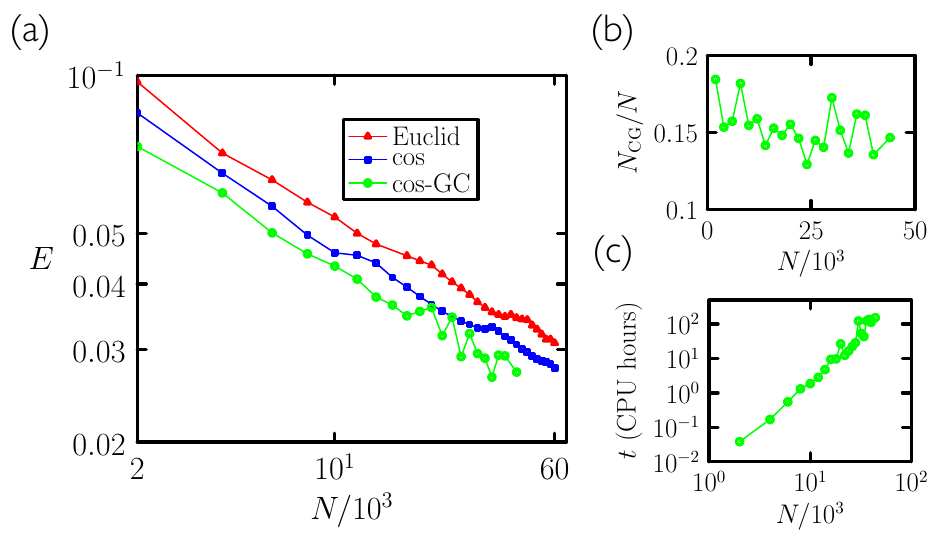} 
   \caption{(a) Error rate $E$ for classification of the $10^4$-digit MNIST test set by nearest-neighbor classification using $N$ symbols from the training set (log-log plot). For red and blue lines the measure of similarity is the Euclidean distance \eq{euclid} and normalized vector dot product \eq{dot}, respectively. The green line was obtained by coarse graining $N$ symbols of the training set into $N_{\rm CG}<N$ memories. Classification accuracy is no worse and is often better upon coarse graining (compare blue and green lines). (b) $N_{\rm CG}/N$ is typically about $15\%$. (c) The CPU time required for coarse graining scales roughly algebraically with $N$.}
   \label{fig1}
\end{figure*}
\begin{figure*} 
   \centering 
   \includegraphics[width=0.8\linewidth]{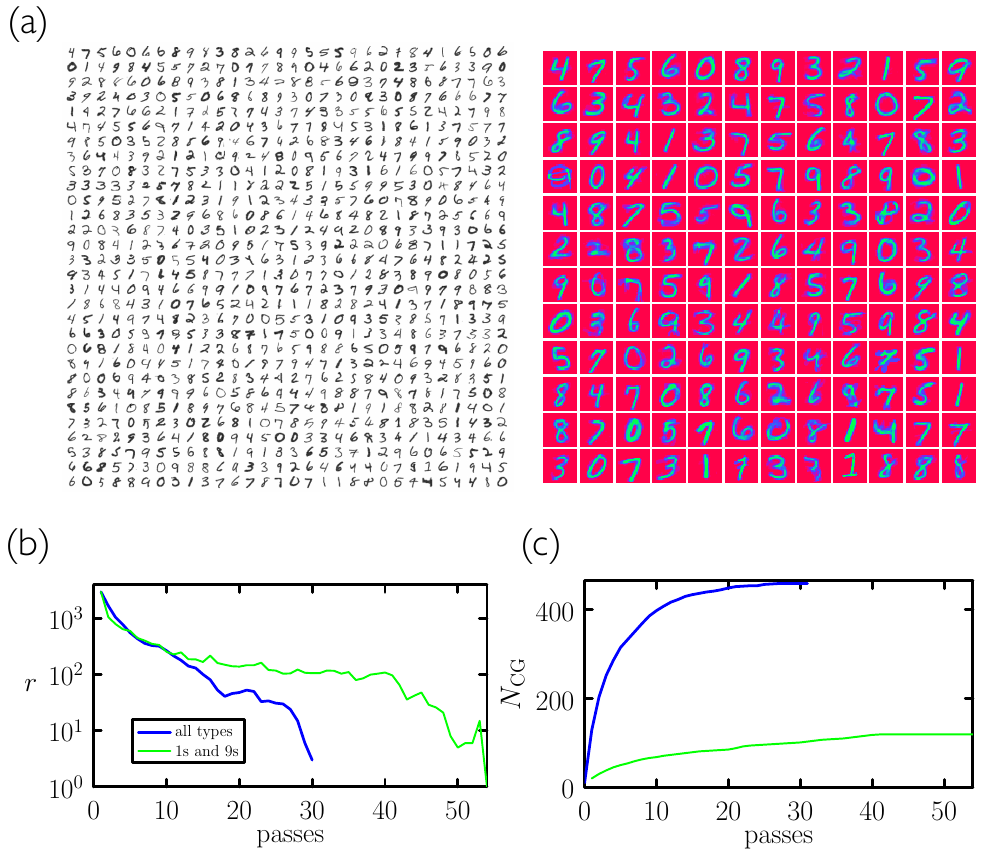} 
   \caption{Coarse-graining algorithm. (a) This 900-symbol batch (left) taken from the MNIST training set yields 144 memories (right) under coarse graining. Both the symbols and the memories achieve a 10\% nearest-neighbor classification error rate on the MNIST test set. (b) Number of rearrangements $r$ required upon each pass through a batch of 3000 symbols taken from the MNIST training set, versus number of passes through the batch. The blue line refers to a batch composed of all types of symbol; the green line refers to a batch composed of 1s and 9s. (c) For the same batches, we show the number of coarse-grained memories produced by the algorithm as a function of the number of passes.}
   \label{fig2}
\end{figure*}

The MNIST data set consists of $L=7 \times 10^4$ handwritten digits, of types 0 to 9 inclusive, divided into a training set of size $6 \times 10^4$ and a test set of size $10^4$. Digits (hereafter called symbols) are grayscale images displayed on a $28 \times 28$ pixel grid, and are represented as $P=784$-dimensional vectors, each entry of which is a scalar between 0 and 255. We divided the value of each entry by 255 to produce a real number between 0 and 1. Let $S^\theta$ be symbol number $\theta$ of the MNIST set, where $\theta \in \{1,\dots, L\}$, and let $S_i^\theta$ be its $i^{\rm th}$ component, where $i \in \{1,\dots, P$\}. Let $T(S^\theta) \in \{0,\dots,9\}$ be the {\em type} of symbol $S^\theta$.
 
In \f{fig1}(a) we show the error rate, the number of incorrect assignments divided by the test-set size of $10^4$, that results from nearest-neighbor classification: we compare each test-set symbol with each of the first $N$ symbols of the training set, and assign to each test-set symbol the type of the most similar training set symbol. A standard measure of similarity is the Euclidean distance between vectors $A$ and $B$, 
\beq
\label{euclid}
D(A,B) = P^{-1}\sum_{i=1}^P (A_i - B_i)^2,
\eeq
with smaller distances representing greater similarity. This measure results in the red line shown in \f{fig1}(a). Another measure of similarity is the normalized vector dot product, the cosine of the angle between vectors $A$ and $B$, 
\beq
\label{dot}
 \hat{A} \cdot \hat{B}=  \frac{\sum_{i=1}^P A_i B_i}{\sqrt{\sum_{i=1}^P A_i^2}\sqrt{\sum_{i=1}^P B_i^2}},
\eeq
with larger values indicating greater similarity. This measure results in the blue line shown in the figure. We shall refer to the normalized vector dot product as the {\em overlap} of vectors $A$ and $B$, 
\beq
\label{ol}
\o(A,B) = \hat{A} \cdot \hat{B}.
\eeq

There exist more discriminating measures of the similarity of two images\c{simard1993efficient,belongie2002shape}. Here our aim is to show how the accuracy of a classifier, using a simple measure of similarity, can be improved by sampling. 

The error rate resulting from nearest-neighbor classification, using the normalized vector dot product as a measure of similarity, is about 2.8\%. This is not close to the error rates produced by the most accurate methods\c{scores} ($\approx0.2 \%$), but neither is it terrible: it is comparable to the rates produced by some of the neural networks of the late 1990s\c{lecun1998gradient}. However, the error rate shown in \f{fig1}(a) decreases approximately algebraically with $N$ (the plot is log-log), and so it is clear that substantial additional reduction in error rate would require orders of magnitude more symbols in the training set. 

Here we show that reduction in error rate by nearest-neighbor classification can be achieved instead by stochastic sampling of the existing training set. The heart of this approach is the coarse-graining algorithm described below. This algorithm turns a subset of training-set symbols into a lesser number of coarse-grained {\em memories}, in such a way that each member of the training subset is correctly classified, via nearest-neighbor classification, by the set of coarse-grained memories.

\subsection{Coarse graining training-set symbols}
\label{cgsec}
We aim to {\em coarse grain} $N$ symbols $S^\theta$ to produce $N_{\rm CG}\leq N$ {\em memories} $M^\alpha$, $\alpha \in \{1,\dots N_{\rm CG}\}$. Each memory has entries $M_i^\alpha$, $i \in \{1,\dots,P\}$, and is of type $T(M^\alpha) \in \{0,1,\dots,9\}$. Memories are linear combinations (centroids) of symbols. To facilitate the addition and removal of symbols from memories it is convenient to write 
\beq M^\alpha = \hat{M}^\alpha/N^\alpha,
\eeq 
where $\hat{M}^\alpha$ is the un-normalized version of memory $\alpha$, and $N^\alpha$ the number of symbols comprising memory $\alpha$. Our coarse-graining strategy proceeds as follows.

\begin{enumerate}
\item Consider a {\em batch}, an ordered collection of $B$ symbols $S^\theta$, $\theta \in \{ 1,\dots, B\}$, taken from the training set. Create a new memory $M^\alpha$ that is equal to the first symbol $\theta=1$ from this batch and is of the same type:
\beq
\label{first}
\hat{M}_i^1 = S_i^1\, (\forall i); \quad N^1= 1; \quad T(M^1) = T(S^1).
\eeq
Record the fact that symbol 1 is now stored in memory 1. Create one memory for each additional type of symbol in the batch (with each memory consisting of one symbol), giving up to ten initial memories. Record the memory in which each symbol is stored (here and subsequently).

\item Return to the start of the batch of symbols and pass through the batch in order. For each symbol, compute its {\em virtual overlap} with all existing memories. If the symbol $S^\theta$ and memory $M^\alpha$ are of the same type, and if the symbol is {\em not} currently stored in that memory, then the virtual overlap is $\o(S^\theta,M^{\alpha+\theta})$ [see \eqq{ol}]. Here $M^{\alpha+\theta}$ is the vector that results if symbol $S^\theta$ is added to memory $M^{\alpha}$; it has components $M_i^{\alpha+\theta} = (\hat{M}_i^{\alpha}+S_i^\theta)/(N^\alpha +1)$. Otherwise (i.e. if the symbol is currently stored in the memory, or if symbol and memory are of distinct type) the virtual overlap is $\o(S^\theta,M^\alpha)$.

\item Let the largest virtual overlap between symbol $S^\theta$ and any memory be with memory $M^\beta$. 
\begin{enumerate}
\item If $S^\theta$ is currently stored in $M^\beta$ then no action is necessary. 
\item If $S^\theta$ is not currently stored in $M^\beta$, and $M^\beta$ is of the same type as $S^\theta$, then add $S^\theta$ to $M^\beta$:
\beq
\label{add}
\hat{M}_i^\beta \to \hat{M}_i^\beta+S_i^\theta; \quad N^\beta \to N^\beta+1.
\eeq
If $S^\theta$ is currently stored in a different memory $M^\alpha$ then remove $S^\theta$ from $M^\alpha$:
\beq
\label{sub}
\hat{M}_i^\alpha \to \hat{M}_i^\alpha-S_i^\theta; \quad N^\alpha \to N^\alpha-1.
\eeq
If $S^\theta$ is not yet stored in a memory then \eq{sub} is not necessary.
\item If $S^\theta$ and $M^\beta$ are of different type then $S^\theta$ has been misclassified. Create a new memory $M^\gamma$ equal to the symbol $S^\theta$ and of the same type:
\beq
\hat{M}_i^\gamma = S_i^\theta; \quad N^\gamma = 1; \quad T(M^\gamma) = T(S^\theta).
\eeq
If $S^\theta$ is currently stored in a different memory $M^\alpha$ then remove $S^\theta$ from $M^\alpha$ [per \eqq{sub}]. If $S^\theta$ is not yet stored in a memory then step \eq{sub} is not necessary.
\end{enumerate}
\item Continue until we have considered all symbols in the batch. Return to 2 and pass through the batch in order again. Note the number of memory increases or symbol relocations that occur on each pass. If the number of each is zero then the algorithm is finished; if not, return to 2 and pass through the batch in order again.
\end{enumerate}
\begin{figure} 
   \centering
   \includegraphics[width=\linewidth]{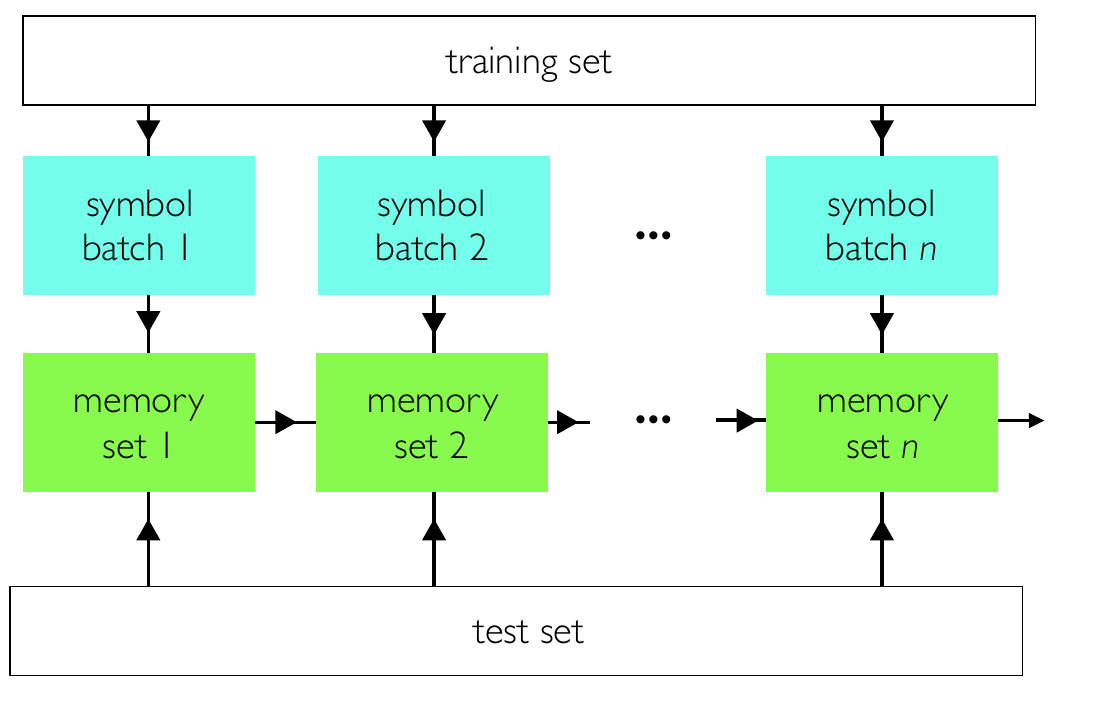} 
   \caption{Coarse graining and sampling. Batches of symbols are constructed by drawing symbols from the training set. Each batch is coarse grained to produce a set of memories. Each set of memories is used to classify the entire test set, and results from all batches are compared.}
   \label{fig3}
\end{figure}
\begin{figure*} 
   \centering
   \includegraphics[width=\linewidth]{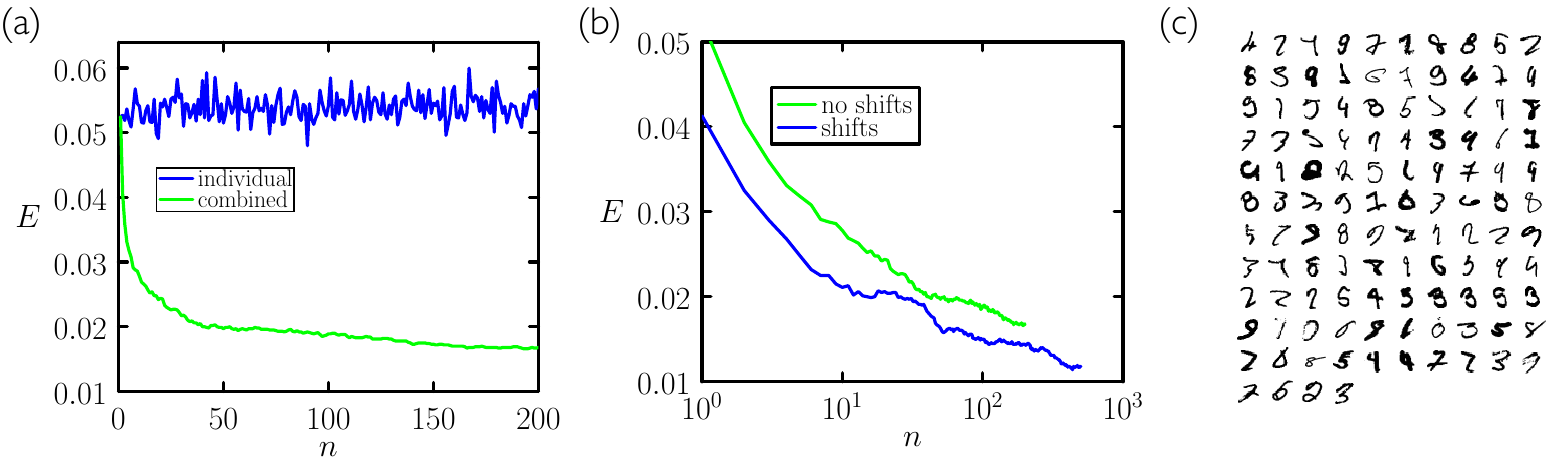} 
   \caption{Sampling. (a) Error rate $E$ achieved on the MNIST test set using $n$ memory sets in parallel (see \f{fig3}). Each set (which contains about 750 memories) was obtained by coarse graining batches of 5000 symbols drawn from the MNIST training set. No single memory set does better than about 5\% error rate on the MNIST test set (blue line). However, in combination they are much more potent (green line). 10 sets (about 7500 memories in total) scores 2.8\%, equal to the rate achieved using all $6\times10^4$ unaltered training-set symbols (see \f{fig1}). With about 200 sets the error rate falls to 1.6\%. (b) Further improvement is possible with better measures of image similarity: the green line is reproduced from panel (a), while the blue line was obtained using linear shifts of symbols. (c) The memory sets used to produce panel (b) together classify the MNIST test at 1.14\% error rate; we show here the 114 misclassified symbols.}
   \label{fig4}
\end{figure*}

This algorithm produces a set of memories that correctly classifies, by nearest-neighbor classification, each digit of the batch of symbols from which it is made. All symbols from the batch are stored in a memory, and some memories contain many symbols. For a wide range of batch types and sizes we observed the algorithm to converge~\footnote{Occasionally we observed two symbols of the same type to shuffle repeatedly between two memories of the same type; at this point the algorithm can be terminated with no loss of accuracy.}. With batches made from the MNIST training set we observed a compression rate of about 6 or 7: the number of coarse-grained memories $N_{\rm CG}$ produced from a batch of $N$ symbols is typically $\approx 0.15N$ [\f{fig1}(b)]. The CPU time taken for the algorithm to run scales roughly algebraically with $N$ [\f{fig1}(c)]. 

We made batches by drawing symbols without replacement from the training set, with frequencies designed to produce, on average, equal numbers of symbol types within the batch (symbol types in the training set are not equally numerous). We picked at random a symbol $S^\theta$ from the training set. With probability $x_-/x_\alpha$ we moved that symbol to the batch. Here $\alpha = T(S^\theta)$ is the type of symbol $S^\theta$; $x_\alpha$ is the number of symbols of type $\alpha \in \{0,\dots,9\}$ in the training set prior to the move; and $x_- \equiv \min_\alpha x_\alpha$. If the move was successful then we reduced $x_\alpha$ by 1 and removed $S^\theta$ from the training set. We then chose another symbol from the training set, and repeated the procedure until a given batch size was obtained. (When constructing many batches for use in parallel we allow each batch to draw from the entire training set independently, without replacement.) 

Significantly, sets of coarse-grained memories are no less accurate than their constituent symbols in classifying the MNIST test set [via nearest-neighbor classification, using \eqq{dot}]: compare the blue and green lines in \f{fig1}. Indeed, in most cases the memories are slightly more accurate than their constituent symbols. In a simple sense the coarse-graining algorithm ``learns'' to partition configuration space, respecting the diversity of the symbol batch and at the same time combining similar-looking symbols. The result is an efficient sampling of that space.

\begin{figure*} []
   \centering
   \includegraphics[width=\linewidth]{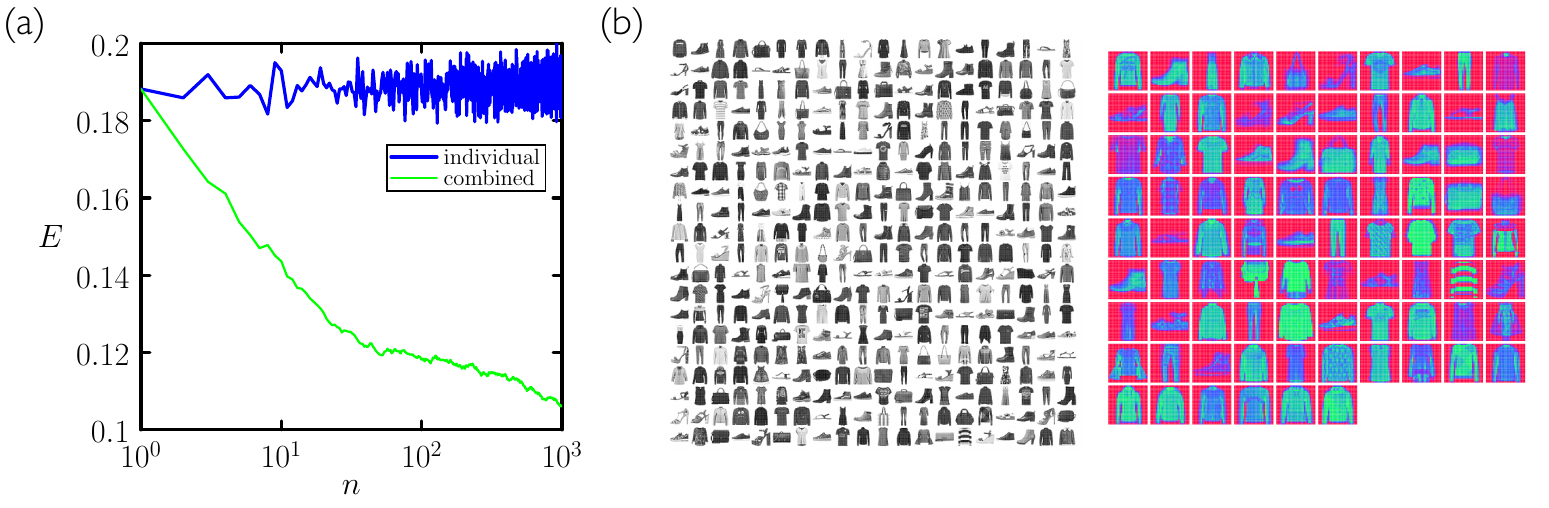} 
   \caption{(a) Similar to \f{fig4}(a), but for the Fashion-MNIST dataset\c{xiao2017fashion}. We use $n$ memory sets each derived from 5000 symbols. For $n=1000$, classification accuracy is 10.5\%. (b) For this data set the symbols-to-memories compression rate is about 4 or 5, rather than 6 or 7 for MNIST, indicating a slightly greater variability per symbol type. This batch of 400 Fashion-MNIST symbols yields 86 memories under coarse-graining.}
   \label{fig4b}
\end{figure*}

In \f{fig2}(a) we show 1000 symbols that yield 144 memories upon coarse graining; both classify the MNIST test set at 10\% error rate~\footnote{Coarse graining the memories themselves results in a set of 36 new memories that achieves a 14\% classification error rate. Repeated coarse graining eventually produces 10 memories (1 per symbol type) and a classification error rate of about 50\%.}. In panels (b) and (c) we show the number of rearrangements made by the coarse-graining algorithm during each pass through a batch of 3000 images, and the total number of memories as a function of the number of passes.
\begin{figure}[] 
   \centering
   \includegraphics[width=\linewidth]{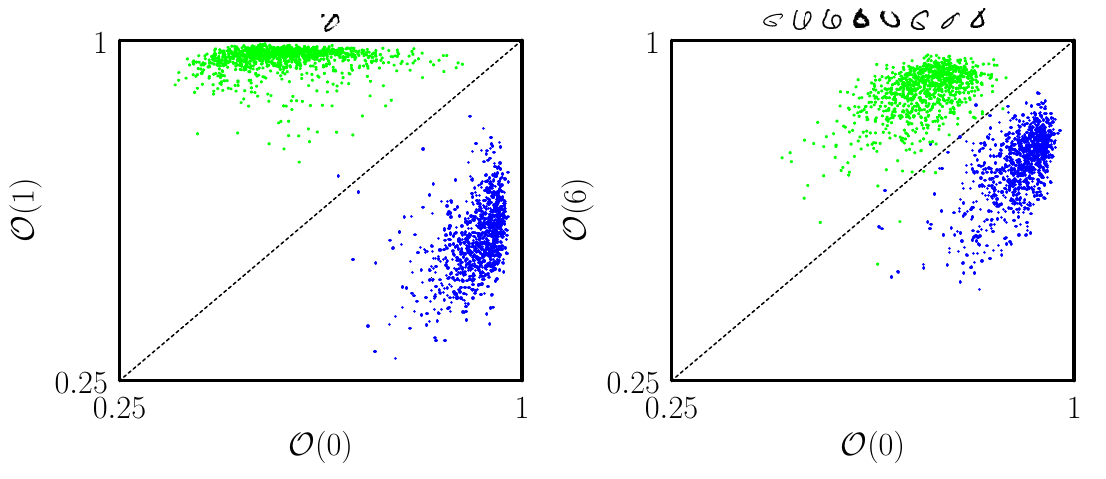} 
   \caption{2 batches of 5000 symbols, of type 0 and 1 or 0 and 6, are coarse grained and used to classify the $(0,1)$- or $(0,6)$-MNIST test set, respectively. Horizontal and vertical axes show the largest overlap between a test-set symbol and the relevant memory type (symbols of type 0 are shown blue). Misclassified symbols are shown at the top of each plot.}
   \label{fig5}
\end{figure}

\subsection{Sampling}
Thus coarse graining achieves a significant compression of information with no loss (and often some slight gain) of nearest-neighbor classification accuracy with unseen data. Used in this way the algorithm can be regarded as a computer memory-saving measure, similar to prototype-identification methods\c{garcia2012prototype} such as the condensed nearest-neighbor approach\c{hart1968condensed,angiulli2005fast}, or to adaptive versions of learning vector quantization\c{nova2014review,fritzke1995growing,marsland2002self}. Because memories are objects not present in the original test set, memory sets can be used in combination to classify unseen data more accurately than can the original training set. If batches are made by drawing stochastically from the training set then each batch is in general different, and will produce different memories upon coarse graining. Such variation can be regarded as a simple means of ``sampling'', in that memories in each set cover different portions of configuration space. This idea is related to that described in Ref.\c{skalak1997prototype}, in which combinations of sub-sampled data are used to improve the accuracy of a nearest-neighbor classifier by constructing voting blocs. Here, coarse graining provides a means of accessing regions of configuration space not present in the original test set.

\f{fig3} shows a simple way in which memory sets can be combined. $n$ batches are built by drawing symbols stochastically from the training set, and each set is coarse grained to produce a set of memories. Each set of memories is used to classify the entire test set by nearest-neighbor classification. By comparing all memory sets we assign to each test-set symbol the type of memory with which it has largest overlap. This scheme is naturally carried out using $n$ processors in parallel.

In \f{fig4}(a) we show the error rate achieved on the MNIST test set using $n$ memory sets in parallel, per \f{fig3}. $n$ batches of $5000$ symbols are drawn from the training set, in the manner described in \s{cgsec}. Each batch is coarse grained, producing $n$ sets of about 750 memories (coarse graining 5000 memories takes about 5 minutes on single 3.1 GHz Intel Core i7 processor). No {\em single} memory set does better than about 5\% error rate on the MNIST test set (blue line). However, used in combination (taking the closest match between a test-set symbol and any memory from the $n$ batches) they are much more accurate (green line). 10 memory sets (about 7500 memories in total) scores 2.8\%, equal to the rate achieved using all $6\times10^4$ unaltered training-set symbols (see \f{fig1}). With about 200 memory sets the error rate falls to 1.6\%. 

Further improvement is possible using larger $n$ or by invoking additional knowledge about the concept being classified: in panel (b) we show data for which memories, once obtained, were compared with test-set symbols by translating memories up to $\pm 3$ lattice sites in either direction. The lowest error rate shown is 1.17\%. When the memory sets used to produce the data in panel (b) are combined, they classify the MNIST test set at 1.14\% error rate. Panel (c) shows the misclassified symbols: some are clearly recognizable, and might be correctly classified if e.g. symbol rotation was accounted for, while others are hard to interpret and would be ``correctly'' classified only if the training set contained a similar symbol of that type.

The error rate is not a strictly decreasing function of the number of batches $n$: adding more batches can increase the error rate if the new batches contain a memory of the wrong type that resembles a test-set symbol more closely than any memory of the correct type. However, the overall trend is that increasing $n$ reduces the error rate.

In \f{fig4b} we repeat the calculation of \f{fig4}(a), but now using the Fashion-MNIST dataset\c{xiao2017fashion}. Fashion-MNIST is a more challenging variant of MNIST consisting of items of clothing of 10 types. Coarse-graining results in a symbols-to-memories compression rate of about 4 or 5 (as opposed to 6 or 7 for MNIST), and 1000 batches of memories each derived from 5000 symbols yields a test-set error rate of 10.5\%. This result compares favorably with the error rates quoted in \cc{xiao2017fashion}, all of which lie above 10\%: for instance, $K$-nearest-neighbor classifiers using the unaltered training set achieve 14\%--16\% error rate, showing the advantage accrued by statistical sampling of the training set.

 Coarse graining and sampling provide a way of improving the efficiency and accuracy of nearest-neighbor classification over that offered by unaltered training-set data. This approach works best with processors used in parallel, with each used to sample and coarse grain a single batch of symbols. The load on any single processor is then relatively light; in the examples described, each processor needs only to store 5000 symbols drawn from the test set, and subsequently store about 750 (for MNIST) or 1000 (for Fashion-MNIST) memories. Training can be done at the same time on all processors, as can classification, with a final step being a comparison between processors of their results.

\subsection{Coarse graining as a genetic algorithm}

The coarse-graining algorithm used here is a method of clustering, resembling a particle-clustering algorithm\c{wolff1989collective,liu2004rejection,whitelam2011approximating} or a supervised version of the $k$-means algorithm\c{wagstaff2001constrained}. It can also be considered to be a type of genetic algorithm: two parents (a symbol and a memory) produce offspring (a memory) that is retained only if it passes a fitness test (recognizing the parent symbol better than does any other memory). This latter feature suggests that memories can be specialized (made ``fitter'') for particular tasks by varying the environment in which memories are made. In simulations described thus far we have used batches that contain, on average, equal numbers of symbols of all types. In ``harsher'' environments, i.e. batches that contain only symbol types that are easily confused, we speculate that memories must be fitter in order to survive. Some evidence in support of this speculation is shown in \f{fig2}(b): a batch of 3000 symbols containing only 1s and 9s requires more passes of the coarse-graining algorithm than does a similarly-sized batch containing all symbol types, suggesting that the coarse-graining algorithm has to work harder to partition configuration space accurately when only similar symbol types are present~\footnote{This batch of 3000 symbols was coarse grained to produce 119 memories (in 23 seconds on a 3.1 GHz Intel Core i7 processor); this set of memories achieves a nearest-neighbor classification error rate of $0.3\%$ on the $(1,9)$-MNIST test set, misclassifying 7 out of 2144 symbols.}. 

In \f{fig5} we show that memories produced by coarse graining training-set batches that contain only two symbol types can tell apart those two types with reasonable accuracy. Two sets of memories, each coarse grained using 5000 symbols of types 0 and 1 or 0 and 6, achieve an error rate of 1 in 2115 or 8 in 1938 on the (0,1)- or (0,6)-MNIST test set, respectively. 
~
\\
~
\\
~
\section{Conclusions}

Nearest-neighbor classification of test-set data by training-set data is a conceptually simple method of classification\c{zhang2007ml,bhatia2010survey,hart1968condensed,angiulli2005fast,hart1968condensed,angiulli2005fast,nova2014review,fritzke1995growing,marsland2002self}. Given a measure of the similarity of two images, we have shown that simple methods of coarse graining and sampling can be used to achieve a more efficient and more accurate nearest-neighbor classification of test-set data than can the unaltered training set. This process creates new symbols from a training set that are a better match for the test set than any of the original training-set symbols. The approach described here, similar to other sampling strategies\c{skalak1997prototype}, works naturally on parallel processors: the more processors, the more accurate is classification. The approach applied to the MNIST and Fashion-MNIST data sets, using nearest-neighbor classification and the simple vector dot product, compares favorably with other forms of classical machine-learning method\c{xiao2017fashion}. While not as accurate as deep-learning methods\c{fmn,scores}, the present method could be improved upon in at least two ways. First, a better measure of image similarity could be used\c{simard1993efficient,belongie2002shape}, or a better distance metric could be learned\c{weinberger2006distance}. Second, the coarse-graining algorithm constructs the centroids of symbols and memories, but other constructions (e.g. splicing together pieces of symbols) are possible; some of these may enable better sampling of configuration space. More generally, coarse graining and sampling schemes might find application in other settings, such as neural networks that make use of memory-based structures.

\begin{acknowledgments}
I thank Marc Pons Whitelam for many discussions about digit recognition, and thank Isaac Tamblyn and Tess Smidt for comments on the paper. This work was performed at the Molecular Foundry, Lawrence Berkeley National Laboratory, supported by the Office of Science, Office of Basic Energy Sciences, of the U.S. Department of Energy under Contract No. DE-AC02--05CH11231. 
\end{acknowledgments}


%

\end{document}